%% file: arXiv_main.tex
\documentclass[10pt,twocolumn,letterpaper]{article}

\usepackage[pagenumbers]{cvpr} 

\input{preamble}

\definecolor{cvprblue}{rgb}{0.21,0.49,0.74}
\usepackage[pagebackref,breaklinks,colorlinks,allcolors=cvprblue]{hyperref}
\usepackage{multirow}
\usepackage{graphicx}
\usepackage{comment}

\def\paperID{37535} 
\def\confName{CVPR}
\def\confYear{2026}

\title{It's a Matter of Time: \\Three Lessons on Long-Term Motion for Perception}

\author{
Willem Davison \quad Xinyue Hao \quad Laura Sevilla-Lara\\
University of Edinburgh\\
{\tt\small \{w.davison, xhao, l.sevilla\}@ed.ac.uk}
}

\begin{document}

\twocolumn[{%
\renewcommand\twocolumn[1][]{#1}%
\maketitle
\centering

\begin{minipage}[t]{0.19\linewidth}
    \centering
    \includegraphics[width=\linewidth]{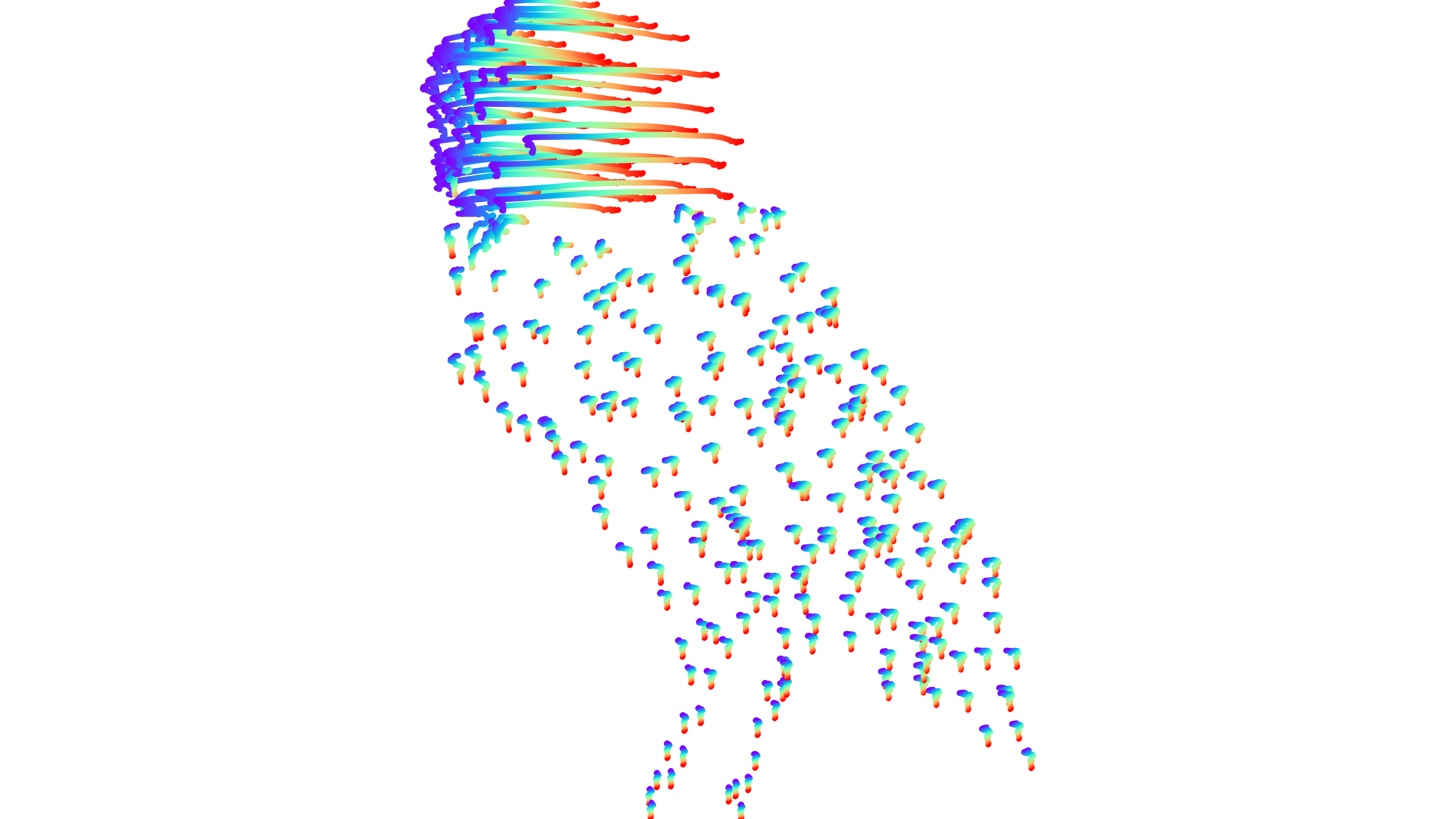}
    (a) Owl
    \label{fig:owl}
\end{minipage}
\begin{minipage}[t]{0.19\linewidth}
    \centering
    \includegraphics[width=\linewidth]{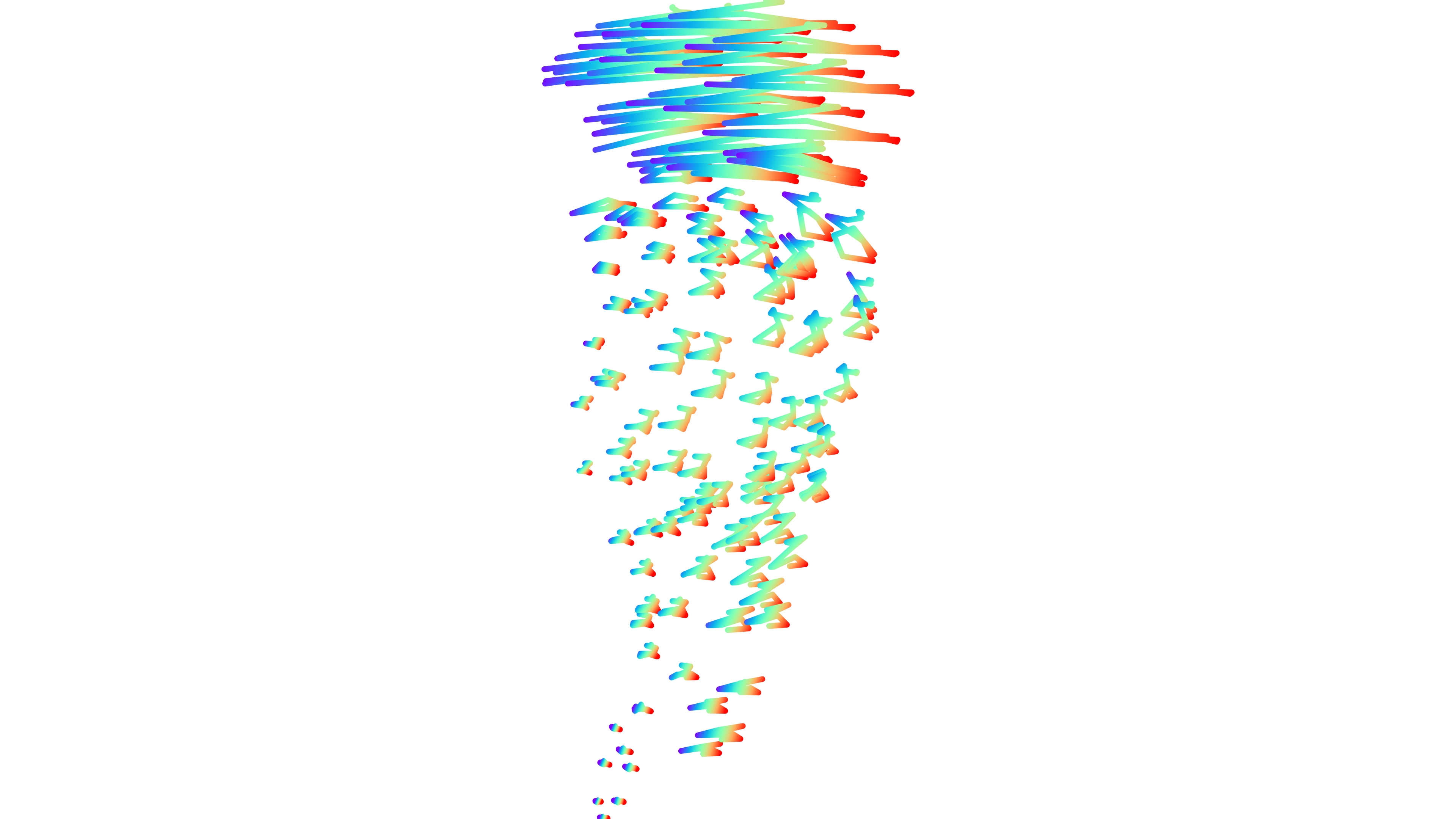}
    (b) Woodpecker
    \label{fig:woodpecker}
\end{minipage}
\begin{minipage}[t]{0.19\linewidth}
    \centering
    \includegraphics[width=\linewidth]{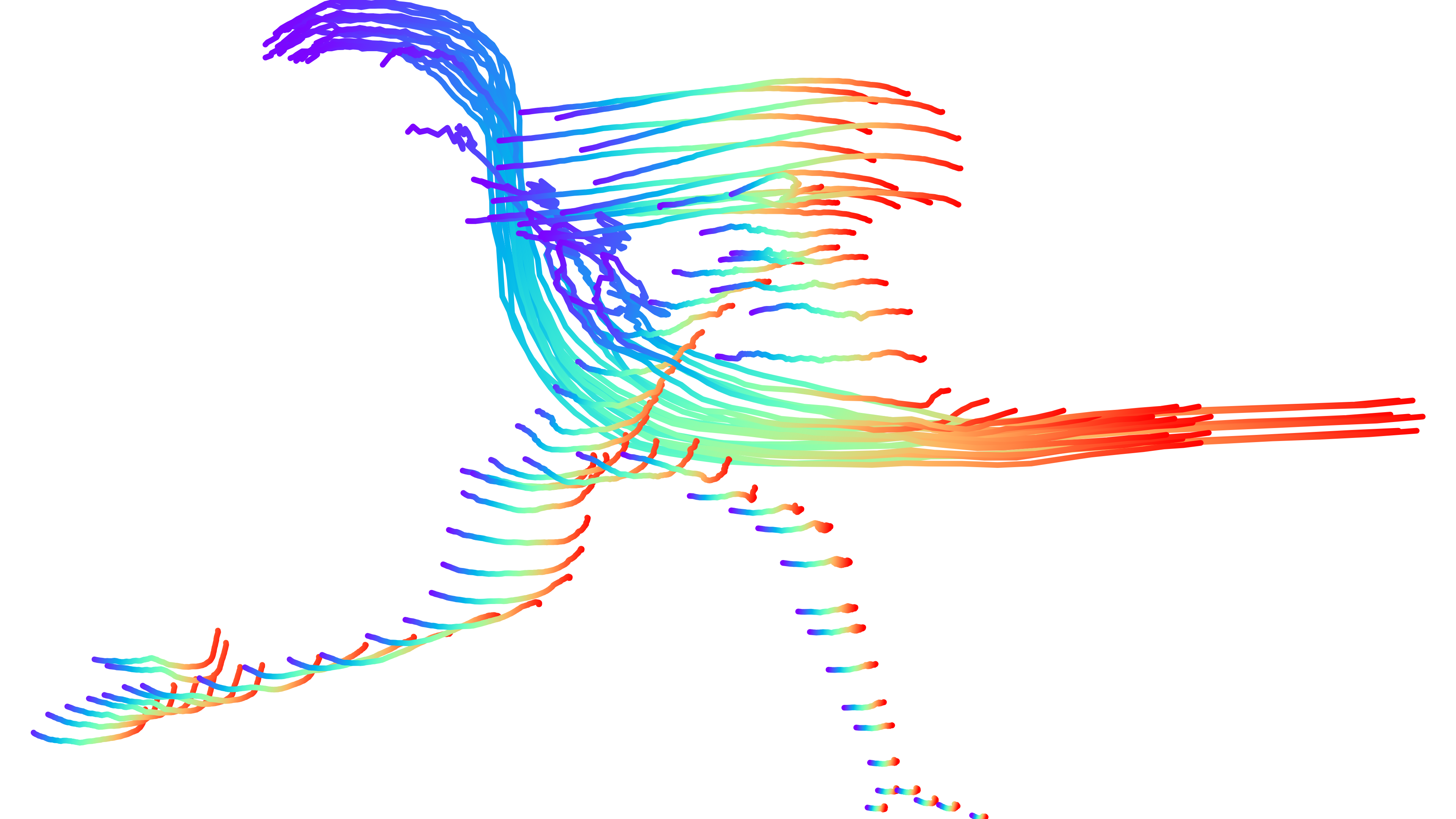}
    (c) Baseball pitch
    \label{fig:baseball}
\end{minipage}
\begin{minipage}[t]{0.19\linewidth}
    \centering
    \includegraphics[width=\linewidth]{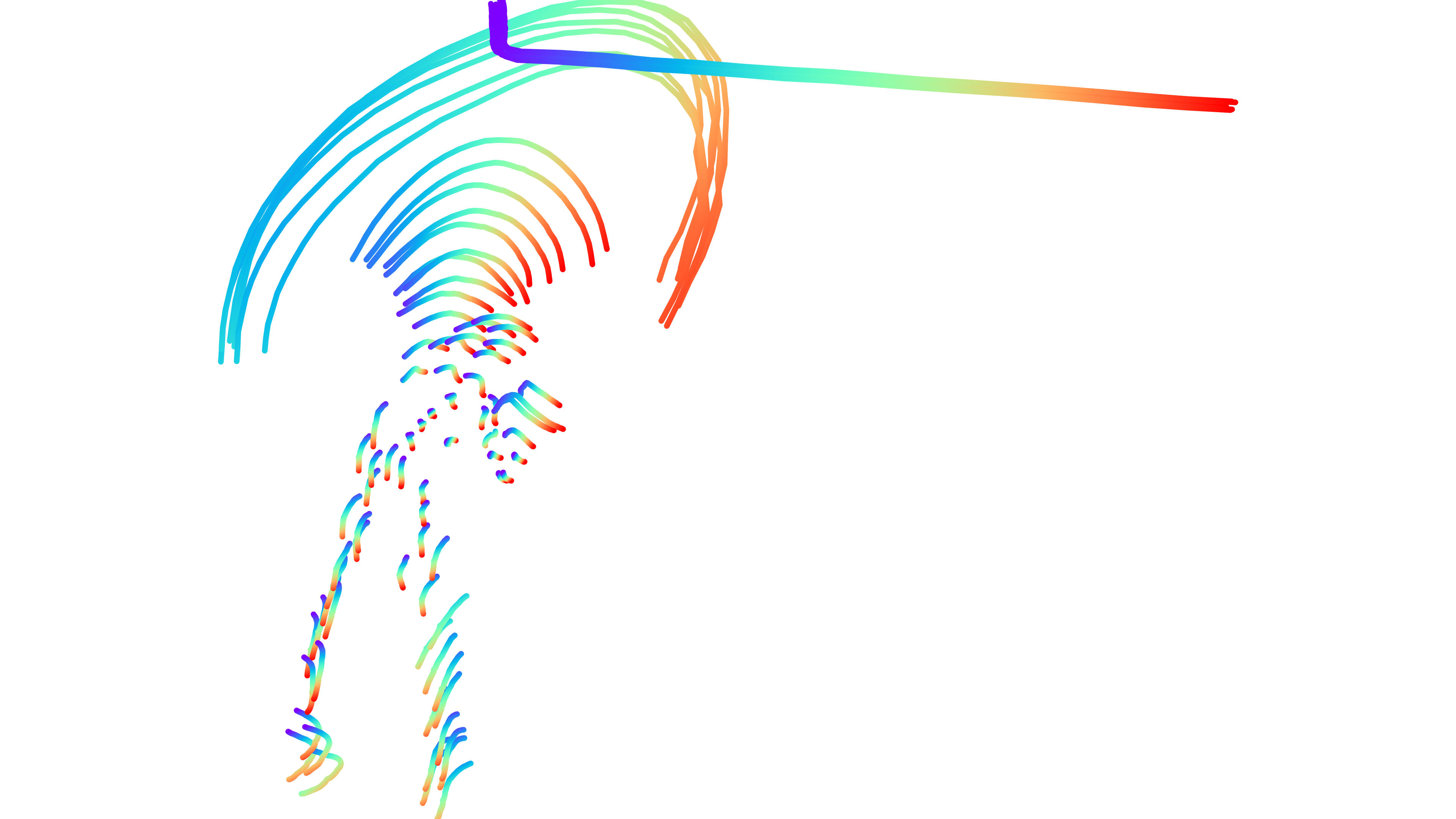}
    (d) Tennis serve
    \label{fig:tennis}
\end{minipage}
\begin{minipage}[t]{0.19\linewidth}
    \centering
    \includegraphics[width=\linewidth]{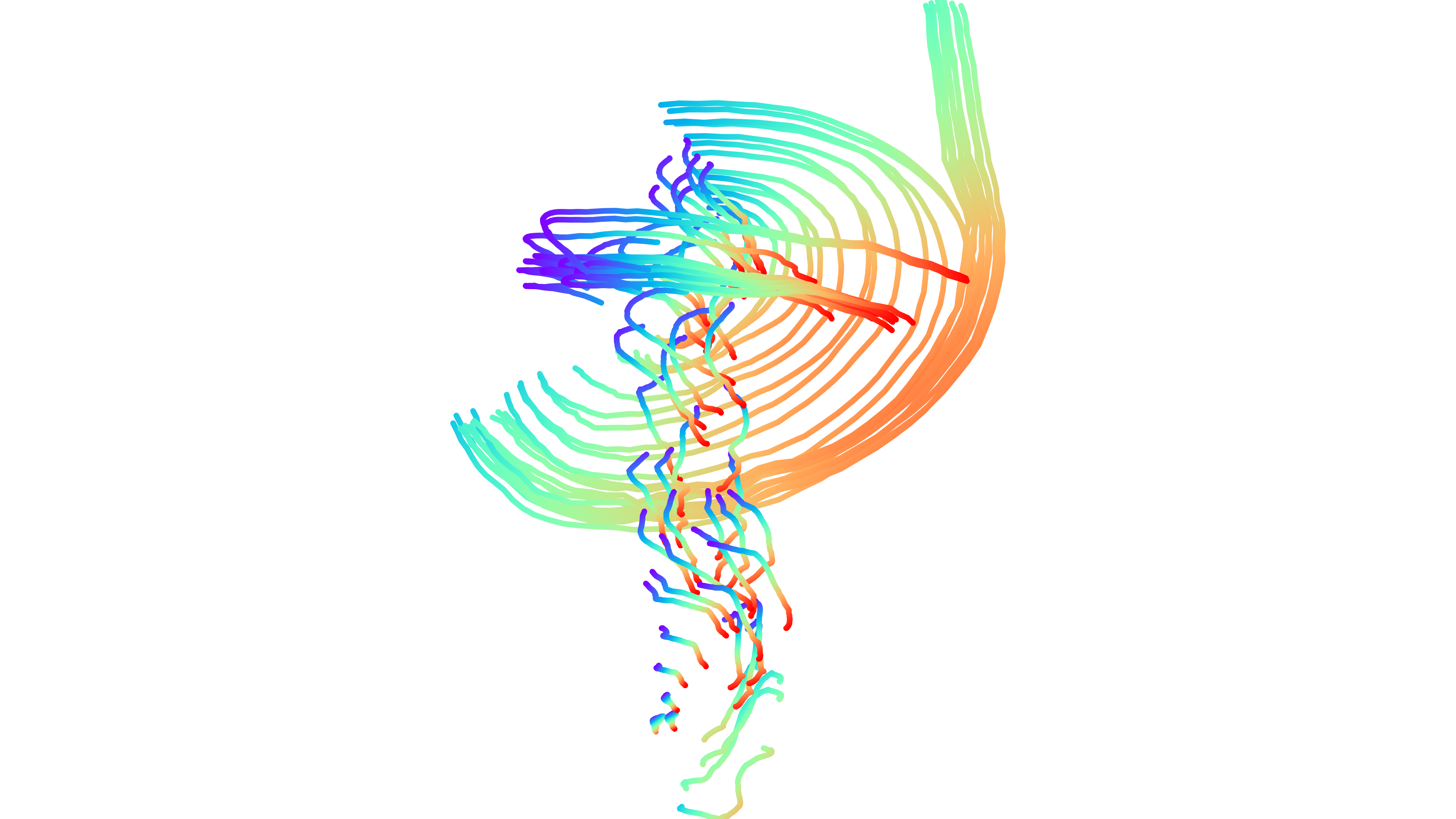}
    (e) Discus throw
    \label{fig:discuss}
\end{minipage}

\captionof{figure}{
    Visualizations of dense point-tracked video instances where color represents temporal evolution (from blue to red). Long-term motion can reveal information about the structure of objects and the actions they are performing. The trajectories alone clearly capture the distinct spatio-temporal dynamics of each action, providing a rich, compact, and effective representation of motion.
}
\label{fig:teaser}
\vspace{1em}
}]

\input{sec/abstract}    
\input{sec/Intro}

\input{sec/RelatedWork}
\input{sec/TechnicalDetails}

\input{sec/Findings}

\input{sec/Conclusion}


{
    \small
    \bibliographystyle{ieeenat_fullname}
    \bibliography{main}
}

\input{sec/X_suppl}

\end{document}

%% file: preamble.tex
\usepackage{graphicx}
\usepackage{caption}
\usepackage{multirow}

\usepackage{float}

\newcommand{\model}{MovT~}
\newcommand{\spmodel}{PixT~}



%% file: sec/abstract.tex
\begin{abstract}
Temporal information has long been considered to be essential for perception. While there is extensive research on the role of image information for perceptual tasks, the role of the temporal dimension remains less well understood: What can we learn about the world from long-term motion information? What properties does long-term motion information have for visual learning? We leverage recent success in point-track estimation, which offers an excellent opportunity to learn temporal representations and experiment on a variety of perceptual tasks. We draw 3 clear lessons: 
1) Long-term motion representations contain information to understand actions, but also objects, materials, and spatial information, often even better than images. 
2) Long-term motion representations generalize far better than image representations in low-data settings and in zero-shot tasks. 
3) The very low dimensionality of motion information makes motion representations a better trade-off between GFLOPs and accuracy than standard video representations, and used together they achieve higher performance than video representations alone. 
We hope these insights will pave the way for the design of future models that leverage the power of long-term motion information for perception. 
\end{abstract}
\vspace{-0.5cm}

%% file: sec/Intro.tex
\section{Introduction}
\label{sec:intro}

Temporal information has long been recognized as fundamental to perception. Takeo Kanade famously summarized the field’s three main challenges~\cite{Wang2019LearningCF} as “correspondences, correspondences, correspondences.” Decades of research in optical flow~\cite{HORN1981185,Black1996TheRobust}, motion capture~\cite{Dyer1995MoCap}, trajectory estimation~\cite{Wang2013}, and point-tracking~\cite{karaev2024cotracker3, doersch2023tapir} had the ultimate goal of leveraging motion as a signal for higher-level perceptual tasks. These tasks have focused largely on action recognition~\cite{Soomro2012UCF101AD,goyal2017something} but have included other tasks such as sign language~\cite{Albanie2021bobsl} and lip reading~\cite{motionbook} to name a few.  

Recent video understanding methods, however, abandoned the goal of using explicit motion information. Instead, they are built with the assumption that temporal information will be learned implicitly in the same way that spatial information is. While this is reasonable, in the context of long-term motion, two challenges arise. The first is memory limitations. Most models can fit around 16-32 frames in memory~\cite{tong2022videomae, wang2022internvideo}, posing the dilemma between sampling frames sparsely (losing short-term information) or sampling densely over a smaller clip (losing long-term information)~\cite{liu-etal-2024-lost}. 
The second challenge is that many tasks do not necessarily force models to learn long-term motion understanding~\cite{SevillaLara2018OnTI}. Self-supervised models~\cite{tong2022videomae, assran2025vjepa2, wang2022internvideo} are trained on tasks that can potentially be solved with short-term temporal information. Large Vision-Language Models (VLMs) also show signs of ``temporal blindness"~\cite{upadhyay2025timeblindness,cores2024tvbench}. In summary, it is not straightforward to establish whether long-term temporal information (even beyond 8 frames) is well captured in current video models. 

Recent work on point-tracking~\cite{karaev2024cotracker3, doersch2023tapir} provides, for the first time, an unparalleled ability to compute accurate, extremely long-term temporal information in real-time. These explicit long-term motion tracks offer an invaluable opportunity for video understanding research to examine the value of long-term information for perceptual tasks.

In this paper, we leverage point-tracking technology as input to learn long-term motion representations. This allows us to ask several long-standing questions: {\em What can be learned about a scene from long-term motion information alone? How does it compare to image information? What properties does long-term motion have in terms of generalization ability and computation cost? }

We test extensively on several high-level perceptual tasks that range from object recognition~\cite{ge2016exploiting} to spatial understanding~\cite{cortes2018advio}, emotion recognition~\cite{RAVDESS}, material properties~\cite{matprop-iccv13}, and action recognition~\cite{goyal2017something,jester}, and measure accuracy, generalization ability, and computational costs, among others. We observe 3 clear lessons that we hope will inform future design of models with strong long-term temporal capabilities. 

In summary, we make the following {\bf contributions}: (1) We analyze and compare long-term motion representations to video representations in a variety of perceptual tasks and observe that motion alone can perform extremely well across all perceptual tasks, even outperforming image recognition, while maintaining very low-dimensionality, (2) We test the generalization abilities of long-term motion representations and demonstrate that they are far more resilient than video representations in two scenarios: when reducing the amount of training data, and when testing on a zero-shot task, (3) Given the low-dimensionality of motion representations, we explore their strength as an efficient encoder of the video. We examine the relationship between accuracy and GFLOPs in motion representations and video representations, as well as the combination of both. We observe clearly that motion representations can add a large amount of information at a very small computational price, paving the way for new research in the combination of video and long-term motion representations.

%% file: sec/RelatedWork.tex
\section{Related Work}
\label{sec:related_work}

\paragraph{Motion and Perception} The role of motion for perception has been studied for decades in different settings and disciplines. The seminal work of Heider and Simmel~\cite{heider1944experimental} in the field of psychology shows us that complex concepts (e.g., joy, threat, fear, anger, escape) can be conveyed simply by motion with extremely simple object information. The widely known Johansson experiment~\cite{Johansson1973} also shows that with the motion of 12 points, humans can recognize complex actions, as long as these points are placed on specific locations of the body, such as joints. In the context of computer vision, there is a wealth of studies~\cite{motionbook} on solving tasks such as action recognition or lip reading using motion. Due to the technological constraints of the time, these models had to either use optical flow, which is general-purpose but short-term, or use motion capture, which is long-term but limited to human motion. Compared to this literature, the recent development of point-tracking technology allows us to use arbitrary objects while maintaining long-term information. In addition, using transformer models~\cite{vaswani2017attention} allows us to capture the information across point-tracks, overall providing an excellent opportunity to address the long-standing question of what motion information can tell us about the world. 
\vspace{-0.3cm}

\paragraph{Video Models with Explicit Temporal Information}

Early video models focused on the task of action classification and often leveraged explicit motion. In the pre-deep learning literature, the IDT model~\cite{Wang2013} showed that hand-crafted features from optical flow and trajectory information were very useful for classification. In the deep learning era, the two-stream paradigm pioneered by Simonyan and Zisserman~\cite{Simonyan2014TwoStreamCN} led to a vast set of work that used one stream where the input was a series of images and another stream where the input was a series of optical flow fields. The high computational cost of optical flow remained a bottleneck, motivating work on using a different motion modality, free of computation: motion vectors from the compressed video~\cite{wu2018coviar}. While motion information was useful, it became clear that dense, sub-pixel accurate motion was not completely necessary~\cite{SevillaLara2018OnTI}, as when an optical flow model was trained together with an action recognition model, the motion features would deviate from the original optical flow. In addition, the arrival of large-scale video datasets such as Kinetics~\cite{Carreira_2017_CVPR}, which was one order of magnitude larger than the predecessors of UCF101~\cite{Soomro2012UCF101AD} and HMDB~\cite{Kuhne:ICCV:2011}, made researchers turn towards an integrated form of learning temporal information implicitly, in the same way that spatial information was. 

\vspace{-0.3cm}
\paragraph{Video Models with Implicit Temporal Information}

More recent work in video understanding has instead aimed to learn to represent temporal information directly from the data in an end-to-end manner, much like models learn to represent spatial information. 

Some of the initial work in the convolutional neural network era learned 3D filters~\cite{Tran2015C3D, tran2018a, Carreira_2017_CVPR}, which meant to substitute completely the use of explicitly computed motion information, such as optical flow. These models incorporated a 3D convolutional filter, which was aimed at capturing temporal short-term structure. The newer era of transformer architectures lent itself particularly well to learning temporal information, as the relation across frames was explicitly modeled~\cite{gberta_2021_ICML, Arnab2021ViViTAV}. These architectures pushed the performance of state-of-the-art significantly. In all these architectures, it is important to note that the quality of the temporal representation will depend on how present that is in the task that the models are trained on. While some tasks do require understanding temporal information~\cite{goyal2017something}, some can often be solved using scene and object cues~\cite{Carreira_2017_CVPR}, which poses a challenge for representation learning in the temporal domain. 

\vspace{-0.3cm}
\paragraph{Benchmarks for Understanding of Temporal Information}

The question of whether a model benefits from using temporal information is naturally intertwined with the choice of task. Some tasks can be solved without much temporal or motion information (for example, when the background of a scene contains contextual clues from which an action may be inferred). Some works have sought to produce benchmarks that not only encourage understanding to stem from temporal and motion information, but require it~\cite{goyal2017something, Sevilla-Lara_2021_WACV, hong2025motionbench}. 
In the context of VLMs, several novel benchmarks also aim at testing the ability to understand temporal information. Causal reasoning benchmarks~\cite{xiao2021next,li2022representation,patraucean2023perception} have sought to push models beyond simple, high-level reasoning. These benchmarks evaluate causal reasoning by introducing questions that require linking temporally distinct events. For example, models must reason about why an action occurred or what might happen if it didn't. Fine-grained temporal benchmarks~\cite{cai2024temporalbench,chen2024rextime} have separately been designed to specifically evaluate a model's understanding of temporal dynamics, such as event ordering, action frequency, and long-range dependencies.

While all these benchmarks are designed to evaluate the ability of a video model to capture motion information, in this work, we wish to expand beyond the tasks where we expect motion to be instrumental. These additional tasks include object recognition and material recognition. 

%% file: sec/TechnicalDetails.tex


\section{Architectures}

We now describe the architectures that we will use for experimentation in this work, using both long-term motion as input, as well as the original pixel input. 

\subsection{Long-Term Motion Architecture}

\begin{figure}[t]
    \centering
    \includegraphics[width=\linewidth]{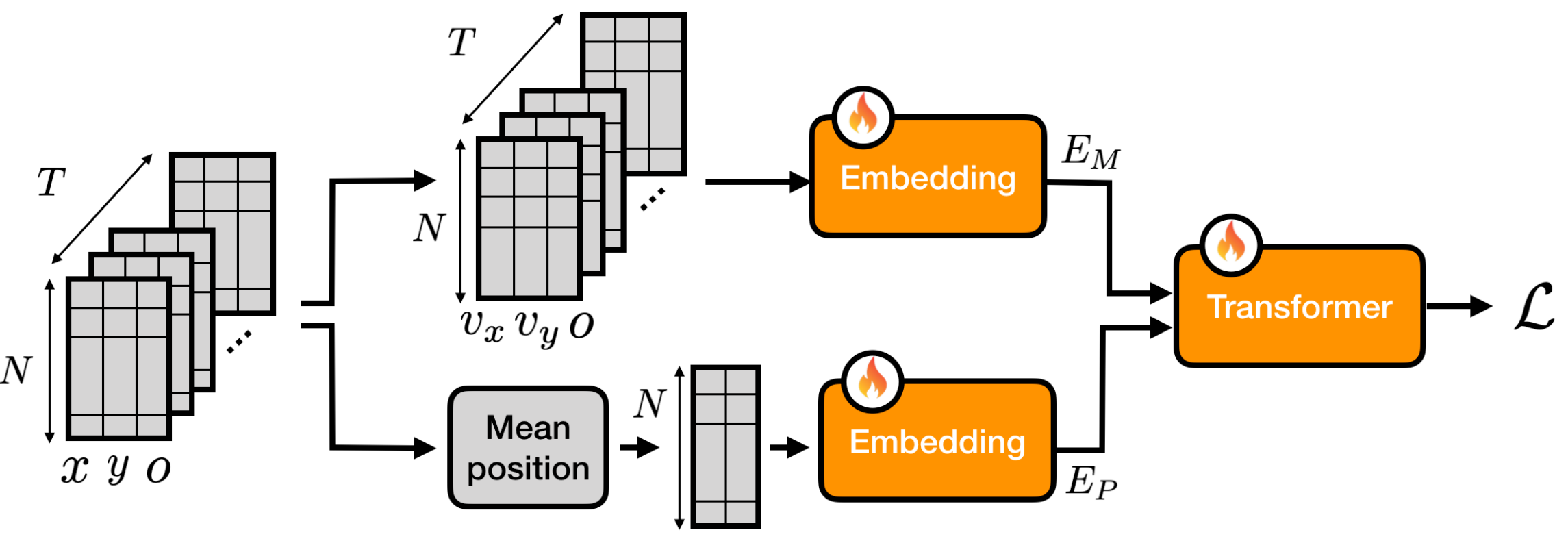}
    \caption{\model Architecture. Point-Tracks are factorized into motion and positional components for embedding. The resulting embeddings are fused for the final spatial transformer layer.} 
    \label{fig:PLACEHOLDERtransformer_model}
\end{figure}

We design a transformer-based architecture~\cite{Arnab2021ViViTAV} that takes as input the point-tracks of a video and outputs a representation that can be used for different tasks. We call this model {\em \model}, for moving point transformer, and it is shown in Figure~\ref{fig:PLACEHOLDERtransformer_model}. Our goal is to design an architecture that will learn long-term motion representations and will allow us to explore the information present in these long-term motion tracks. Therefore, it is a fairly standard architecture without ``bells and whistles" not designed to outperform other models in benchmarks, but as a tool for scientific investigation. 


\paragraph{Input Point-Tracks}

We use CoTracker3~\cite{karaev2024cotracker3} to estimate the point-tracks. Each point-track may be represented as a series of 2D coordinates along with a third, binary dimension, occlusion. Let $P$ be the set of $N$ input point-tracks along $T$ frames, it can be described as:  
\begin{equation}
    P\in\mathbb{R}^{T\times N\times 3}. 
\end{equation}


\paragraph{Motion Encoder}
We compute the velocity of each point at each time step by taking the temporal difference of the position coordinates given by the input tensor, which yields two values $v_x, v_y$. This is concatenated with the occlusion channel, which results in a tensor of dimensions $[N, T, 3]$.
We use a small network to learn an embedding for this relative velocity. This network is comprised of a multi-layer-perception (MLP), followed by a 1D convolutional layer and a temporal max-pooling. This yields a vector for each point-track, which captures its motion, which we call $E_M$, as it is an embedding ($E$) of the motion ($M$) of the track. $E_M$ is of size $1 \times D$. All the motion embeddings are stacked together in a tensor of size $[N, D]$. 

\vspace{-0.2cm}

\paragraph{Position Encoder}
To provide spatial context to the point-tracks, we calculate a positional embedding. For each point-track, we compute the average location of the $x$ and $y$ values over time. This creates a single coordinate representing each track's average location, resulting in a tensor of shape $[N, 2]$. This is input to an MLP to compute $E_{P}$, for the embedding of the position, and is also of size $1 \times D$. We stack all positional embeddings together into an $[N, D]$ tensor. 

\vspace{-0.2cm}
\paragraph{Cross-Point Transformer}

The $E_{M}$ and $E_{P}$ embeddings are concatenated and input to a transformer. 
So far, the representations were learned over each individual track. We now use a transformer to learn information across point-tracks. We use a standard 4-layer transformer encoder followed by a mean-pooling over the $N$ dimension, which outputs a representation of the video $E_{F}$, for final. 
For the classification tasks, we use a standard cross-entropy loss. For the regression tasks, we use a standard mean-square-error.

\subsection{Image-based Architectures}


To maintain a fair comparison, the model used in our pixel-input baseline experiments is almost identical to \model with two key differences: 1) We do not calculate ``velocity" and instead use raw RGB as the input features and 2) we use a static positional embedding which learns from the structure of the RGB input. We call this architecture \spmodel for pixel transformer. Experimenting with \model and \spmodel allows us to keep constant all factors, including: model capacity, pre-training, fine-tuning, etc. 

In addition, we also wish to explore the value of motion representations in relation to state-of-the-art video models. For this, we use the standard VideoMAE~\cite{tong2022videomae}, which contains more parameters than the simpler models, and has been extensively pre-trained. Leveraging this model will allow us to inspect the added value of motion representations in a different setting. We do this by employing a simple late fusion architecture between \model and VideoMAE.

\section{Datasets}\label{sec:datasets}

The choice of tasks and datasets in our experiments is of particular importance. We aim to have a broad representation of perceptual tasks, as we want to understand the advantages and limitations of each of the two streams of information (spatial/images, temporal/point-tracks). The tasks should also require both spatial and temporal information to solve them. We examine 5 different tasks: action recognition, object recognition, emotion recognition, material property recognition, and spatial understanding. 

\vspace{-0.2cm}
\paragraph{Action Recognition - SSV2 and Jester}

Both the Something-Something-V2 (SSV2)~\cite{goyal2017something} and Jester~\cite{jester} action recognition datasets are designed for human action recognition. SSV2 specifically focuses on context and object-independent classification and Jester comprises hand gesture classes, which are also, by nature, context and object-independent. In our experiments, we use a subset of the SSV2 dataset defined by the ``Temporal Dataset"~\cite{Sevilla-Lara_2021_WACV}. This is done to ensure that the dataset used comprises only classes for which temporal information has been proven to be essential. We preprocess both datasets by sampling videos down to 32 frames in length.


\vspace{-0.2cm}
\paragraph{Object Recognition - VB100}

To test the effectiveness of motion signatures in the context of object recognition, we use the Video Bird Dataset (VB100)~\cite{ge2016exploiting}. This dataset consists of videos spanning 100 bird species classes. Video instances are first standardized to an fps of 30 and a resolution of 640$\times$360 before being clipped into 90-frame (3-second) clips.

\vspace{-0.2cm}
\paragraph{Emotion Recognition - RAVDESS}

The RAVDESS dataset~\cite{RAVDESS} contains head-shot videos of actors displaying 7 distinct emotions. Employing MediaPipe, we select 60 facial landmark points for tracking and the static, initial positions of our points are used to evaluate a baseline performance. The subsequent movement of the points is used to classify the displayed emotion of each subject in each clip.

\vspace{-0.2cm}
\paragraph{Material Properties - MITFabric}

We wish to investigate how well sparse motion may capture material deformation and how well this deformation can be used to infer a material's properties. We chose the MITFabric dataset~\cite{matprop-iccv13}, which comprises videos of 30 materials being subjected to three wind speeds. Videos are first separated into 300-frame (10-second) long clips with a stride of 150 frames. 
We predict the two material properties in this dataset: stiffness and area weight. The performance is then evaluated by calculating Pearson's regression coefficient between predicted and ground truth values for both material properties.

\vspace{-0.2cm}
\paragraph{Spatial Understanding - ADVIO}

We use the Authentic Dataset for Visual-Inertial Odometry (ADVIO)~\cite{cortes2018advio} to assess spatial understanding. This dataset comprises first-person video sequences with corresponding ground truth translation and orientation pose data. In our experiments, video sequences are sampled into 120-frame (2-second) clips, and a 20$\times$20 uniform grid of points is sampled on the first frame of each sequence. We train a regression model to predict the ground truth pose data and performance is evaluated by calculating both the translational and rotational relative-pose-errors (RPE). Averaging over all time-step predictions, we then report the root-mean-squared-error (RMSE) translational RPE and RMSE rotational RPE as our final evaluation metrics.

\begin{figure*}[htbp]
    \centering
    \includegraphics[width=\linewidth]{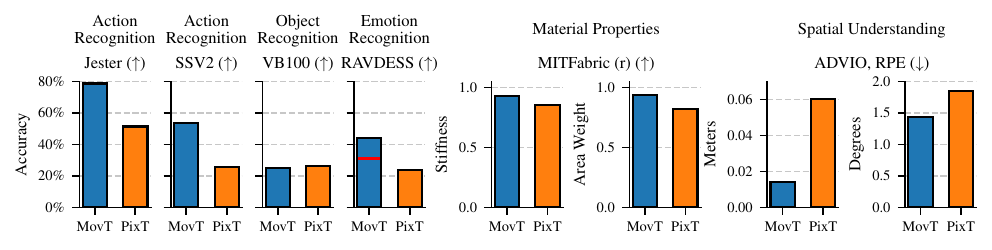}
    \caption{{\bf Lesson 1:} Motion representations can solve a variety of tasks with high accuracy, comparable or better than video representations. We display the task-relevant performance metric in both the point-track and pixel input studies. Up and down arrows $(\uparrow/\downarrow)$ signify whether a higher or lower performance metric is better. The red line on RAVDESS MovT at 31\% marks the performance of our baseline experiment, where the facial landmark points are input to MovT for training.}
    \label{fig:motion_compression}
\end{figure*}

\begin{figure*}[htbp]
    \centering
    \includegraphics[width=0.8\linewidth]{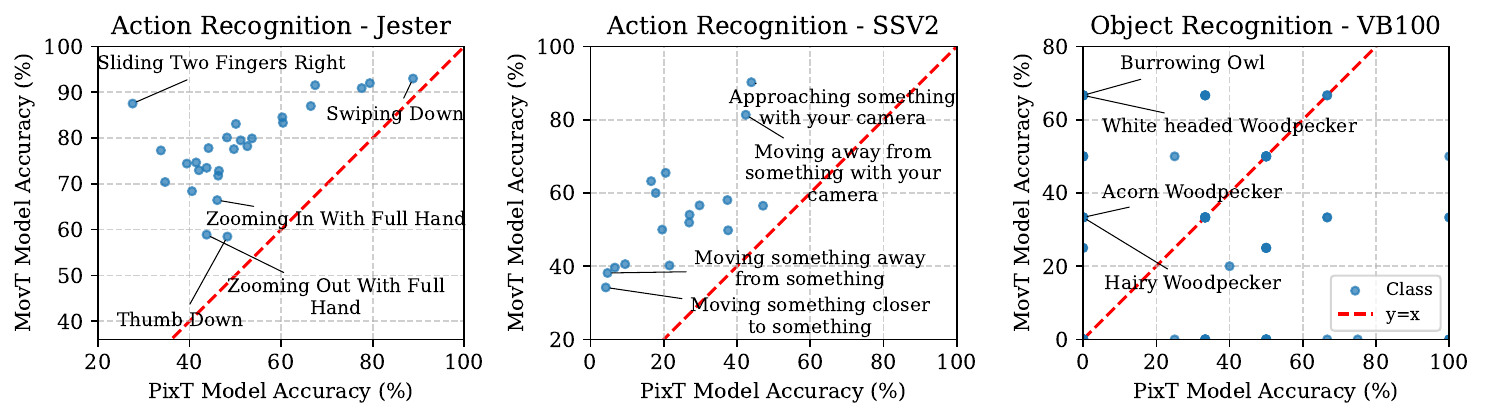}
        \caption{{\bf Lesson 1:} Many classes can be solved with motion information alone, including examples where image representations perform poorly. Scatter plots displaying a comparison of the per-class accuracy of our \model and \spmodel models in three classification tasks.}
    \label{fig:MovTvsPixT_per_class_accuracy}
\end{figure*}

\section{Training Details}

In all \model and \spmodel experiments, our models are implemented using PyTorch. We use the AdamW optimizer and apply gradient clipping with a maximum norm of 1.0 to stabilize training. We use the learning rate scheduler, ReduceLROnPleteau, in all experiments. We use MSE as the loss function for both regression tasks, MITFabric and ADVIO. In those examples that feature multiple clips across each video (MITFabric and VB100), clip probabilities are aggregated to produce a single final prediction at the video level. In all training instances, both point-track and pixel input, we train over 50 epochs and use a starting learning rate of 1e-4, with the exception of the MITFabric training, where a lower starting learning rate of 1e-5 is used. All models are trained over 50 epochs.

For training our VideoMAE fusion models, when the number of input frames is set to 1, the tubelet size is also configured to 1. For inputs with more than one frame, the tubelet size is fixed at 2. We train all models without using any warm-up epochs. For the Jester and VB100 datasets, the learning rate is set to 5e-3 with a total of 30 training epochs, while for the SSV2 dataset, the learning rate is set to 1e-3 and training is performed for 40 epochs.

%% file: sec/Findings.tex


\section*{Lesson 1: \\What Does Motion Reveal About a Scene?}

\paragraph{Question: } We first examine the capabilities of motion representations for high-level perceptual tasks. What can motion representations tell us about the world? How do they compare to image representations? And do they effectively capture long-term temporal information?

\vspace{-0cm}
\paragraph{Experiment Set-up: } In this experiment, we use all tasks and datasets, and train \model and \spmodel for each of them. Image input tends to be of larger dimensionality than point-tracks. To be able to make a fair comparison, we fix the size of the input to be the same for images and point-tracks. We simply use GroundedSAM~\cite{GroundedSAM} to obtain both a bounding box and segmentation around the object. In the \model experiments, the segmentation mask is used to sample points from within. In the \spmodel experiments, the pixels found within the bounding box are used as our model's input. In an ablation study, we investigate the impact of varying the number of input frames to MovT and PixT.

\vspace{-0.2cm}
\paragraph{Results: }

Results are shown in Figure~\ref{fig:motion_compression}. We observe that the long-term motion of \model alone succeeds in all tasks: action recognition, object recognition, spatial understanding, and material recognition. Many of these tasks can actually be solved with higher accuracy than the equivalent video model \spmodel. The largest gaps are in the action recognition datasets as well as in the spatial understanding. It is particularly interesting to observe that even for the object recognition task, knowing how a bird moves is as useful as knowing what it looks like. 

We are also interested in observing the distribution of the classes along the accuracy dimensions for each of the models. In other words, we ask the question: are the hard and easy classes the same for \model and \spmodel? 

For three classification tasks, we compare the per-class accuracy of the two models and show the results in Figure~\ref{fig:MovTvsPixT_per_class_accuracy}. We observe that in the action datasets there is a clear correlation between the difficulty of a class for the motion model and the video model. However, for the object recognition model, we observe that the points are much more scattered, suggesting that indeed some classes have a stronger motion signal and others have a stronger temporal signal. We illustrate some of the outliers, and observe that, for example, owls and woodpeckers, as in Figure~\ref{fig:teaser}, are well recognized by the motion model and not at all by the pixel-input model.

In this ablation study, we perform tracking on the longer videos from SSV2. We then progressively crop all instances in the time dimension, such that we are sampling the middle `N' frames of every sample. This is done for both MovT and PixT, and each model is retrained from scratch for every data point. We find that PixT’s performance stagnates early, while MovT’s continues to increase until around 40 frames, a temporal length significantly exceeding the typical 16-frame memory limit of current SOTA models. Note that in this dataset, 48 frames is nearing the maximum length of all videos. The plateauing of MovT's performance near this length may therefore be attributed to dataset-specific information redundancy (e.g., the first and last few frames of each video will typically involve the actor beginning and ending the recording and so not performing the action). It may be assumed then that this 40-frame plateau is task-specific and that on other datasets with longer videos, temporal information may be captured effectively well beyond this length.

\vspace{-0cm}
\begin{figure}[htbp]
    \centering
\includegraphics[width=0.55\linewidth]{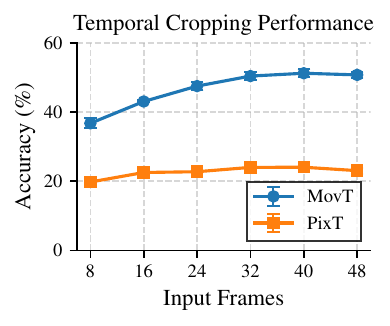}
    \caption{Performance impact of temporal cropping SSV2 videos.}
    \label{fig:temporal_cropping}
\end{figure}
\vspace{-0cm}

\section*{Lesson 2: \\Motion Generalizes Better Than Images}

Let us consider the thought experiment of learning to recognize when an object is falling. In image space, different objects will yield different videos depending on their color, background, material, etc. However, in motion space, the tracks will be much more similar across objects. This gives us the intuition that the space of possible motions might be smaller than the space of possible images, and therefore it could be learned with less data. In addition, Johansson's experiment~\cite{Johansson1973} shows us that motion information can be very largely downsampled while still being discriminative. How can we leverage the low dimensionality of motion information? In this experiment, we propose to examine how this low-dimensionality affects a core issue in visual learning: generalization ability.  

\vspace{-0cm}
\paragraph{Question:} Do motion representations generalize better than video representations? This general question can be further split into two aspects: Can motion representations be learned with {\em less data} than image representations? and, can they perform better on {\em zero-shot} tasks than image representations? 

\vspace{-0cm}
\paragraph{Experiment set-up: }
We split our experiments into two, one for each question. For both of them, we chose the Jester and SSV2 datasets, as they are the largest. 

First, we progressively reduce the training set to 60\%, 40\%, 20\%, and 10\%. We train \model and \spmodel on each of these settings and measure the drop in performance. The lower the drop in performance, the stronger the generalization ability of the model. In a separate experiment, we consider data reduction on a per-instance basis; progressively reducing the spatial fidelity of the training and testing data. Once again, we train \model and \spmodel from scratch each time and measure the drop in performance.

Second, we use both models trained on SSV2 to be tested (using just a single linear layer) on Jester, and compare the performance to that of using a model trained on Jester. 

\vspace{-0cm}
\begin{figure}[htbp]
    \centering
\includegraphics[width=0.98\linewidth]{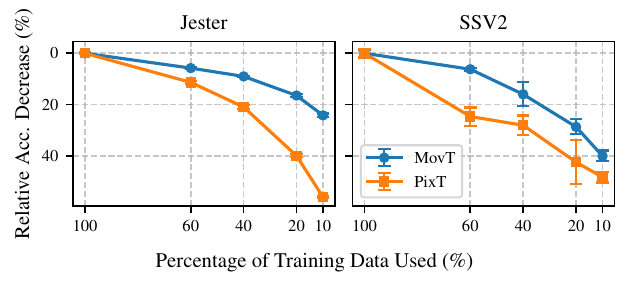}
    \caption{{\bf Lesson 2:} Motion models have stronger generalization abilities in low-data settings. We test the effect of reducing the size of the training set progressively, and measure the resulting accuracy for \model and \spmodel. In both datasets, the \model loses less accuracy than \spmodel model, showing that motion representations indeed generalize better than image representations.}
    \label{fig:generalization}
\end{figure}
\vspace{-0.5cm}
\begin{figure}[htbp]
    \centering
\includegraphics[width=0.8\linewidth]{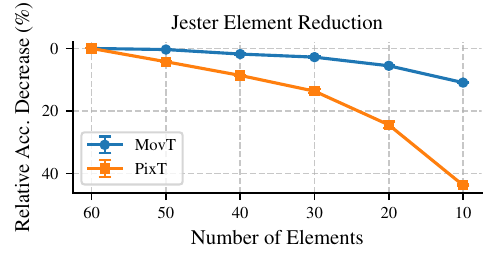}
    \caption{{\bf Lesson 2:} Motion models have stronger generalization abilities in low-data settings. We test the effect of reducing the training data on the Jester dataset on a per-instance basis. We reduce the spatial fidelity of the input instances by randomly downsampling pixels and point-tracks.}
    \label{fig:per-instance_reduction}
\end{figure}
\vspace{-0.5cm}

\begin{figure}[htbp]
    \centering
    
\includegraphics[width=0.55\linewidth]{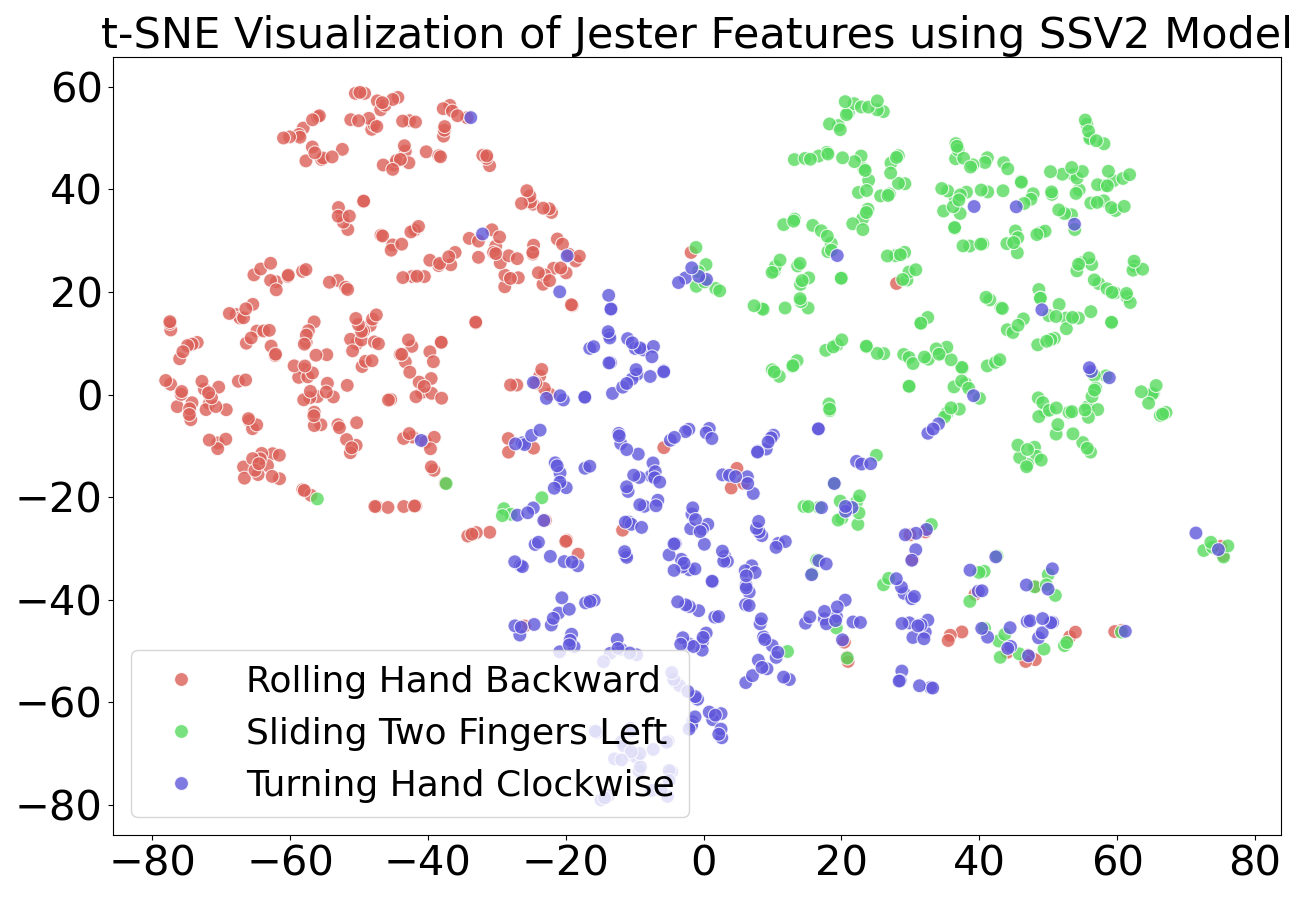}    \includegraphics[width=0.43\linewidth]{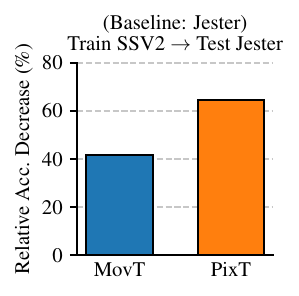}
    \caption{{\bf Lesson 2: } Motion models have stronger generalization abilities in zero-shot tasks. Left: t-SNE visualization of Jester classes using the \model trained on a different dataset, SSV2. We observe that classes are extremely well clustered using the \model representation, even for a zero-shot task on an unseen dataset. Right: Loss of accuracy on the zero-shot task of training on SSV2 and testing on Jester.}
    \vspace{-0.5cm}
\label{fig:generalization2}
\end{figure}

\paragraph{Results:}
Results in Figure~\ref{fig:generalization} show that \model achieves better performance when using only a portion of the dataset in both action recognition tasks, Jester and SSV2. In fact, for the rather extreme case of using 10\% of the data on the Jester dataset, \model suffers a drop of 23\%, while \spmodel suffers a drop of 56\%. 

The result of the zero-shot experiment is shown in Figure~\ref{fig:generalization2} (right). We observe that while \spmodel loses over 60\% of the accuracy when tested on a new, unseen dataset, \model loses around 40\%, showing that indeed long-term motion representations are more general and less dataset-specific than the video counterparts. We also visualize the motion representations using the t-SNE~\cite{maaten2008visualizing} technique in Figure~\ref{fig:generalization2} (left). Each point corresponds to a video from the Jester dataset, which is encoded using the motion representation trained on the SSV2 dataset. We observe that the representation of videos of the same class are well clustered, although the representation was not trained on that dataset. All of these results show a consistent message: that long-term motion demonstrates strong generalization capabilities even in the zero-shot setting.

\input{sec/large_table}

\section*{Lesson 3: Motion Representations Are More Efficient Than Images}

\vspace{-0cm}
\begin{figure}[htbp]
    \centering    \includegraphics[width=0.8\linewidth]{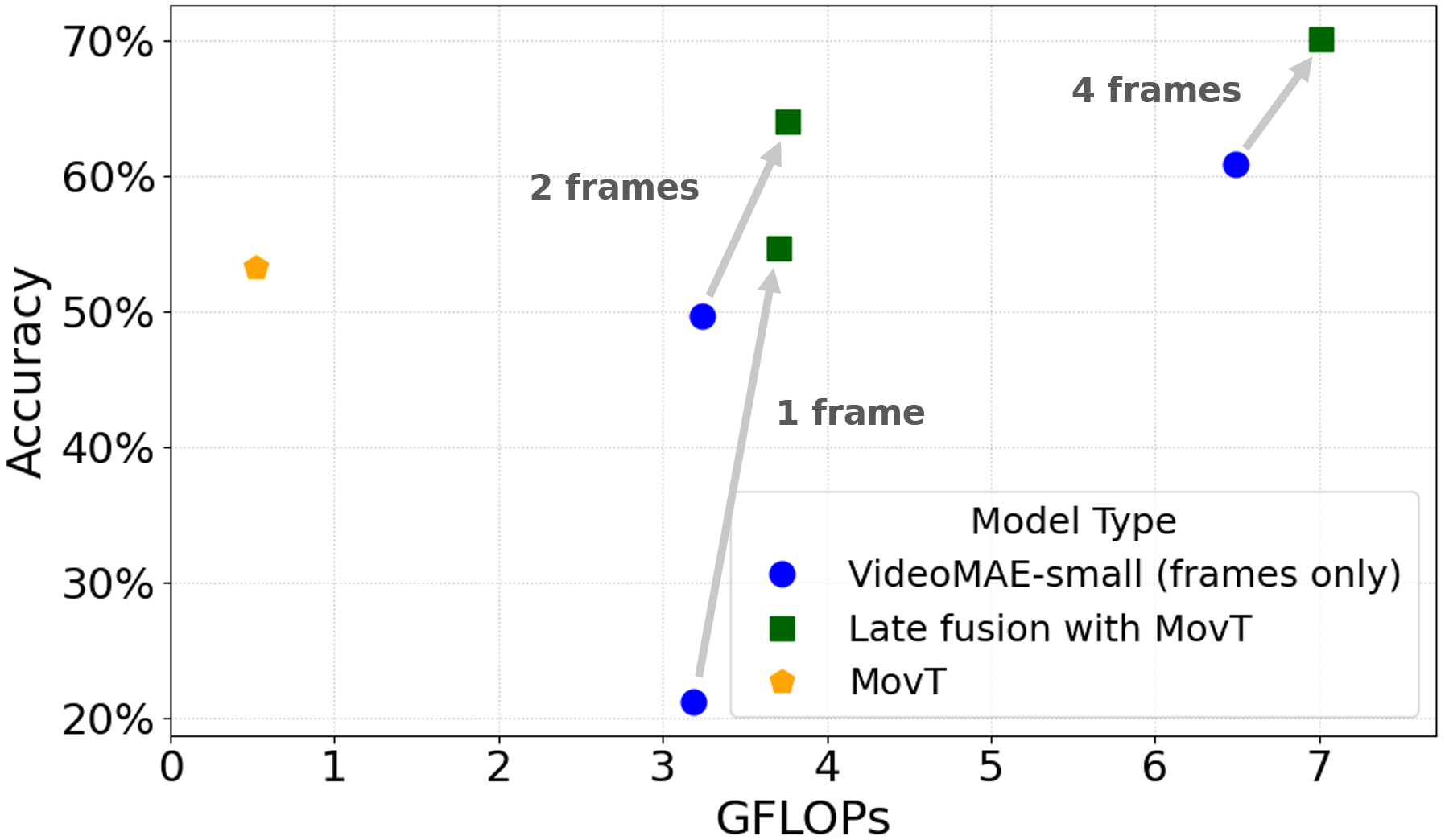}
    \caption{{\bf Lesson 3: }Motion representations can improve accuracy by a large margin while adding a low computational cost. Performance gain from incorporating \model into VideoMAE. We can easily observe that \model offers the best trade-off between computation and accuracy. We also observe that the combined model is the Pareto-front, showing optimality with respect to using only VideoMAE.}
    \vspace{-0cm}
    \label{fig:trade_off_ssv2}
\end{figure}

Video representations are notorious for being computationally expensive as the input dimensionality is very large, and therefore, computational cost is an important aspect to consider when designing models. As mentioned before, point-tracks can have much lower dimensionality, while still maintaining strong performance across tasks. This leads us to consider another aspect to explore in the context of motion representations: the trade-off between compute and accuracy. In this section, we consider this trade-off in the context of motion representations, state-of-the-art video representations, as well as the combination of both. 
The combination is particularly interesting as it may cast light on the design of future low-compute models. The long-term motion representations can be seen as a lightweight ``summary" of a video. Combining them with a few frames from the computationally expensive video could be an optimal compromise between accuracy and computational cost. 

\vspace{-0.3cm}
\paragraph{Questions:} What is the trade-off between accuracy and GFLOPs for motion representations and state-of-the-art video representations? How can we combine state-of-the-art models and long-term motion representations? 

\vspace{-0.3cm}
\paragraph{Experiment Set-up: }

As a representative of state-of-the-art models, we adopt VideoMAE as the video representation and train the model using only raw video frames. While there are more recent models, VideoMAE has been extensively studied, and it has a reasonable computational cost, making it a good choice for experimentation. 

For the combination of both representations, we use late fusion between VideoMAE and \model. We experimented with more sophisticated integrations, including cross-attention, however, late-fusion showed the best performance. The ablation study is in the supplementary material. The fusion is applied at the representation level, where the logits from both branches are linearly combined with equal weights (0.5 and 0.5) before the final classification. 





\vspace{-0cm}
\paragraph{Results:}

Table~\ref{tab:videomae_logit_fusion} presents the performance improvements achieved by integrating point-track information into frame-only models. The results show that datasets requiring strong temporal reasoning, such as Jester and the temporal classes of SSV2, benefit the most, with accuracy gains up to 44\%. Notably, these improvements incur only a marginal increase in computational cost, which is practically negligible. 

To make the results of Table~\ref{tab:videomae_logit_fusion} easier to observe, we visualize the results of SSV2 in Figure~\ref{fig:trade_off_ssv2}. Here, we can easily observe the performance boost on the SSV2 dataset as a representative example following the incorporation of motion cues from point-tracks. 
These observations indicate that point-track representations provide a powerful signal for video understanding, particularly for tasks that depend on temporal reasoning, and can substantially enhance the capacity of image-based models to interpret video content. Further analysis of the efficiency properties of the models is shown in the Supplementary Material.

\begin{figure}[htbp]
    \centering
    \includegraphics[width=\linewidth]{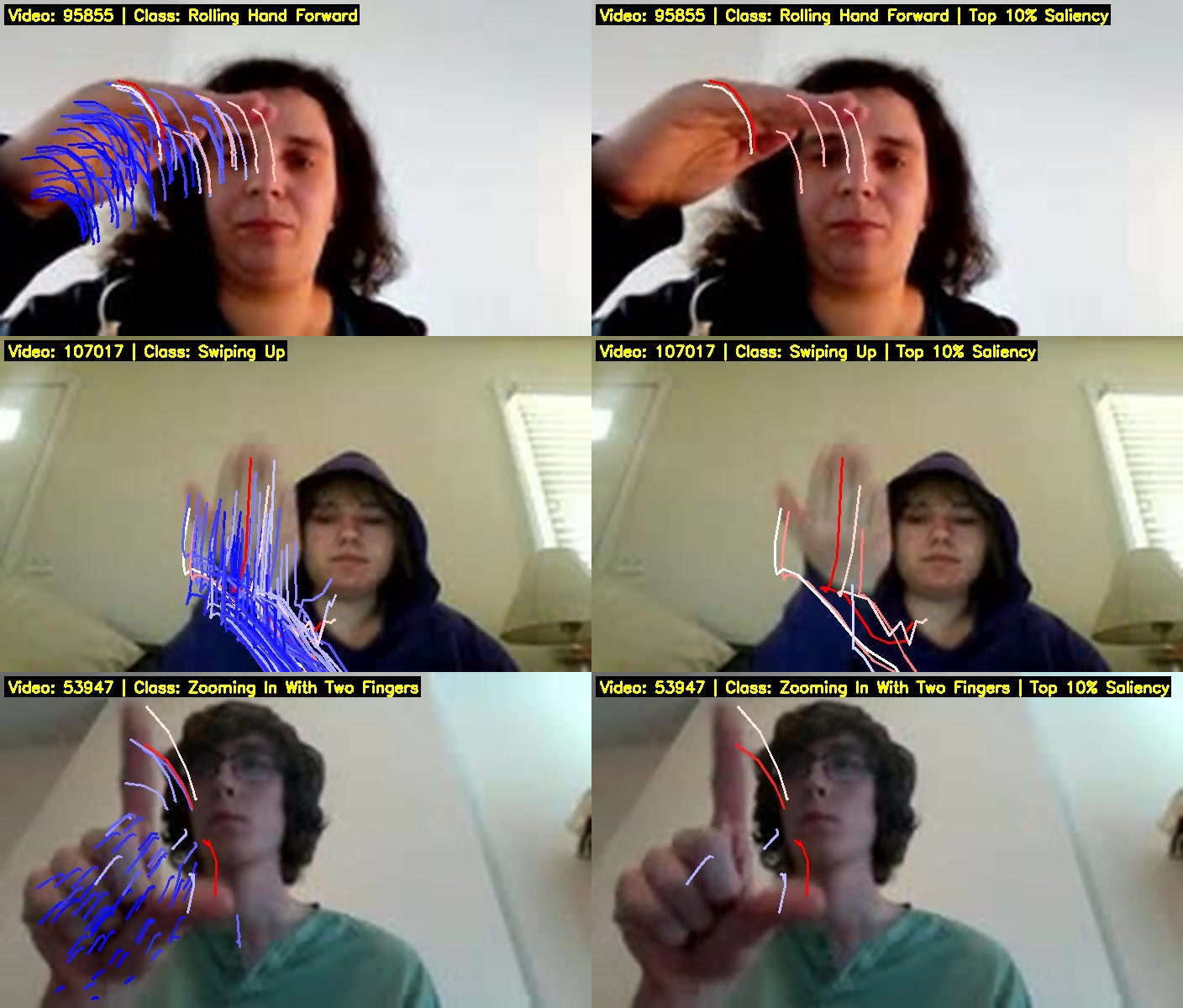}
    \caption{Visualization of point-tracks in the Jester dataset. For each video example, the left part shows all point-tracks, where colors transition from blue to red to indicate increasing importance scores. The right part highlights the top 10\% most important tracks. As shown, the most important point-tracks are mainly located on the skeleton-related regions.}
    \label{fig:important_point_tracks_analysis}
\end{figure}

\section{Interpreting and Scaling Motion Representations}

In our previous experiments, we observed that motion representations are highly effective in solving a wide range of video downstream tasks, such as action recognition, object detection, and spatial understanding. Moreover, motion features exhibit strong generalization capability across various domains and datasets. These observations inspire a further question: \textit{what specific components within motion representations are truly responsible for such versatility?} 

We employ the \model representation as in our previous experiments to conduct this analysis. We adopt an approach inspired by prior works~\cite{simonyan2013deep,selvaraju2017grad,hao2025principles} that utilize gradient-based methods to visualize and interpret which parts of an input contribute more to the final classification. Specifically, we compute the gradients with respect to the model’s output to obtain class saliency maps, assigning each point-track an importance score. Notably, we use the ground-truth label for gradient computation in this analysis. Figure~\ref{fig:important_point_tracks_analysis} visualizes the importance scores with different colors of different point-tracks on the Jester dataset. The color scheme is as follows: red point-tracks are the most discriminative, and blue are the least, with the colors in between signifying intermediate discriminative values. For each of the examples, we see on the left the full set of point-tracks on the hand, and on the right the top 10 most discriminative point-tracks. From the visualization, we observe that the most critical motion components correspond closely to the underlying skeletal structure. Specifically, for the Jester dataset, where most actions are centered around hand movements, point-tracks located on the fingertips, finger joints, and the base of the palm exhibit notably higher importance scores. These results suggest that the model primarily relies on fine-grained skeletal cues to capture discriminative motion patterns in dynamic hand gestures.

\begin{figure}[htbp]
    \centering
\includegraphics[width=0.9\linewidth]{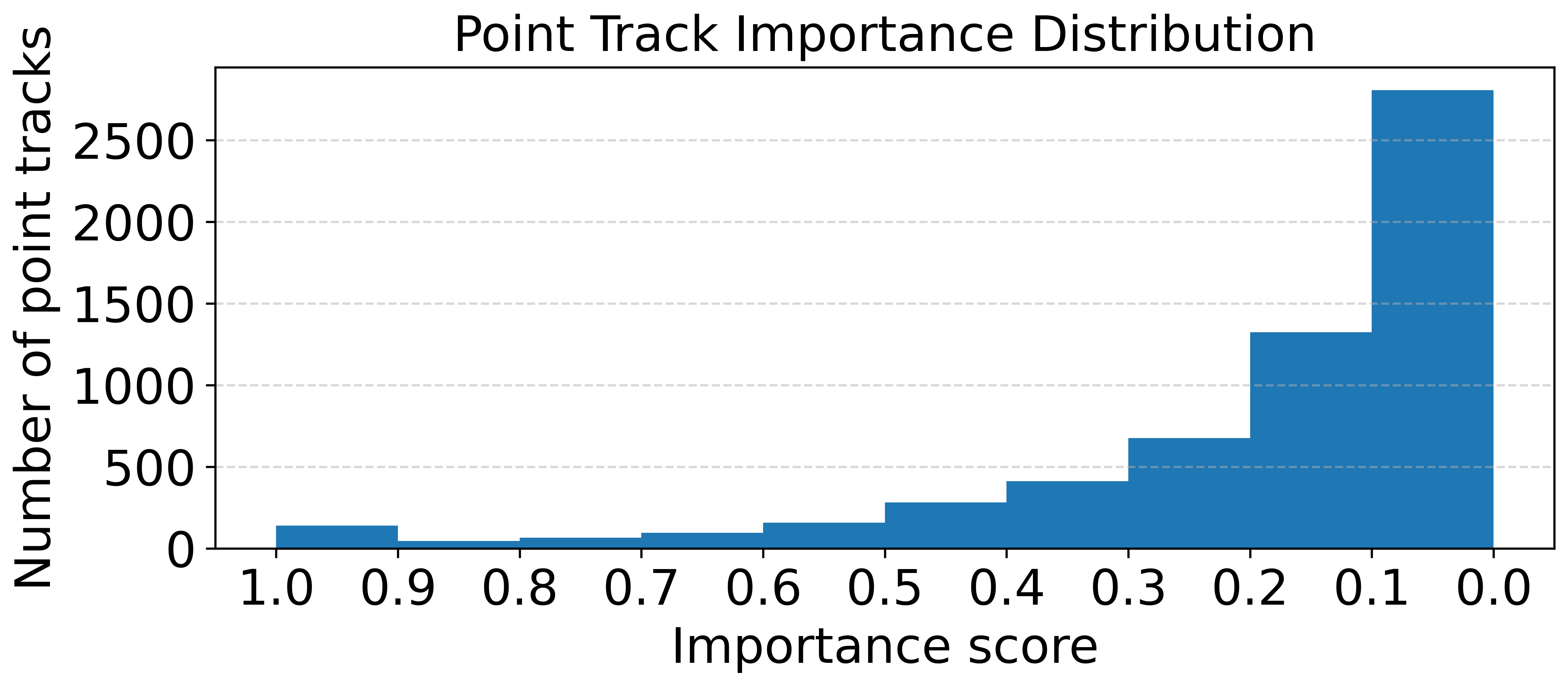}
    \caption{Point-track importance score histogram.
\vspace{-0cm}}
    \label{fig:importance_scores}
\end{figure}

In Figure~\ref{fig:importance_scores}, we show the distribution of point-track importance scores across 100 videos. This demonstrates that a small number of point tracks play a decisive role in inference. On average, each video contains only 5–6 point tracks with scores greater than 0.5. This displays MovT's robustness to point-track downsampling, an effect also shown in Figure~\ref{fig:per-instance_reduction} for randomly downsampling.

%% file: sec/large_table.tex
\begin{table*}[htbp]
  \centering
  \resizebox{\textwidth}{!}{%
  \begin{tabular}{lccccccccc}
    \toprule
    \multirow{2}{*}{Models} & \multicolumn{3}{c}{\textbf{Jester}} & \multicolumn{3}{c}{\textbf{SSV2 (temporal)}} & \multicolumn{3}{c}{\textbf{VB100}} \\
    \cmidrule(lr){2-4} \cmidrule(lr){5-7} \cmidrule(lr){8-10}
    & Frames & Top-1 Accuracy & GFLOPs & Frames & Top-1 Accuracy & GFLOPs & Frames & Top-1 Accuracy & GFLOPs \\
    \midrule
    \midrule
    VideoMAE-small & 1 & 30.49\% & 3.19 & 1 & 21.15\% & 3.19 & 1 & 5.11\% & 3.19 \\
    \textbf{+ \model} & 32 & 74.53\% \textcolor{green}{(+ 44.04\%)} & 3.64 \textcolor{red}{(+ 0.45)} & 32 & 54.68\% \textcolor{green}{(+ 33.53\%)} & 3.71 \textcolor{red}{(+ 0.52)} & 90 & 17.57\% \textcolor{green}{(+ 12.46\%)} & 7.25 \textcolor{red}{(+ 4.06)} \\
    \cmidrule(l){1-10}
    VideoMAE-small & 2 & 75.10\% & 3.24 & 2 & 49.65\% & 3.24 & 2 & 29.13\% & 3.24 \\
    \textbf{+ \model} & 32 & 87.47\% \textcolor{green}{(+ 12.37\%)} & 3.69 \textcolor{red}{(+ 0.45)} & 32 & 64.00\% \textcolor{green}{(+ 14.35\%)} & 3.76 \textcolor{red}{(+ 0.52)} & 90 & 29.88\% \textcolor{green}{(+ 0.75\%)} & 7.30 \textcolor{red}{(+ 4.06)} \\
    \cmidrule(l){1-10}
    VideoMAE-small & 4 & 90.64\% & 6.49 & 4 & 60.91\% & 6.49 & 4 & 33.18\% & 6.49 \\
    \textbf{+ \model} & 32 & 92.14\% \textcolor{green}{(+ 1.50\%)} & 6.94 \textcolor{red}{(+ 0.45)} & 32 & 70.20\% \textcolor{green}{(+ 9.29\%)} & 7.01 \textcolor{red}{(+ 0.52)} & 90 & 30.78\% \textcolor{red}{(- 2.40\%)} & 10.55 \textcolor{red}{(+ 4.06)} \\
    \midrule
    \textbf{\model} & 32 & 70.01\% & 0.45 & 32 & 53.31\% & 0.52 & 90 & 15.92\% & 4.06 \\
    \bottomrule
  \end{tabular}%
  }
  \caption{ {\bf Lesson 3: } Motion representations can improve accuracy by a large margin while adding a low computational cost. Results of the fusion model for VideoMAE and \model on Jester, SSV2, and VB100 datasets. Adding a motion representation improves the accuracy of VideoMAE in most settings by a fairly wide margin of up to 44\%. This particular improvement comes at a moderate cost of 0.45 GFLOPs. The improvement is most noticeable in the case of action recognition classes.}
\label{tab:videomae_logit_fusion}
\end{table*}

%% file: sec/Conclusion.tex
\section{Conclusion}

We provide a comprehensive exploration of the role of sparse long-term motion representations for perception in video understanding. We build on the recent success of point-tracking technology as a building block for this study. We design a thorough set of experiments to compare a long-term motion representation (MovT) to an equivalent video representation (PixT) as well as to a representative state-of-the-art, pretrained model (VideoMAE). 
Our \model and \spmodel comparison studies demonstrate that long-term motion representations possess higher information density than their image representation counterparts and outperform them in almost all tasks. 
We also demonstrate that motion representations exhibit superior generalization properties to image representations. This is displayed both in low-data and zero-shot settings. The low-dimensionality of motion representations is leveraged in a fusion experiment, where the standard VideoMAE is trained in tandem with MovT. We observe a remarkable trade-off between a significant performance increase and a marginal increase in model complexity. This finding is demonstrated across multiple tasks and in settings with varying numbers of images used as input to VideoMAE. Finally, we analyze the importance of individual point-tracks on final classification. We find that a minimal amount of point-tracks can effectively capture the motion of a video and that point-track representations have exceptional robustness against random downsampling.

%% file: sec/X_suppl.tex
\clearpage
\maketitlesupplementary

\section{Comparing Fusion Methods Ablation Study}

We investigate several strategies for fusing a video frame–only model with a point-track model. All experiments employ the ViT architecture and our \model model. We explore multiple designs: (1) adding cross-attention blocks to the ends of both model branches and training the cross-attention module from scratch, (2) freezing the early encoder layers and fine-tuning only the cross-attention components, and (3) introducing cross-attention and jointly fine-tuning all parameters.

\begin{figure}[h!]
    \centering
    \includegraphics[width=\linewidth]{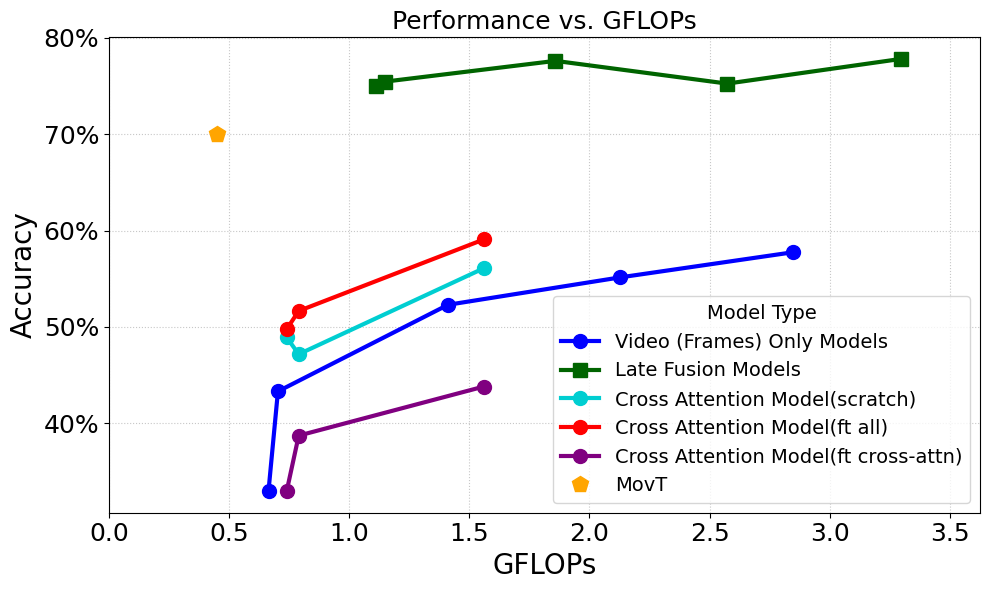}
    \caption{Comparison of different fusion techniques on the Jester dataset.}
    \label{fig:ablation_study_fusion_model}
\end{figure}

Figure~\ref{fig:ablation_study_fusion_model} provides a comparison of the performance of these methods. The experiments show that late fusion yields a substantial performance gain compared to the cross-attention alternatives. We therefore use the late-fusion architecture for all fusion experiments in our final design.

\section{Immediate Implementation}

\begin{figure}[htbp]
    \centering
\includegraphics[width=0.98\linewidth]{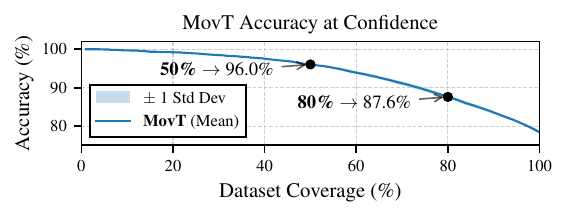}
    \caption{MovT performance on Jester above given confidence thresholds. Over a threshold of 0.978 confidence, we achieve 50\% dataset coverage at 96.0\% accuracy. Over a threshold of 0.743 confidence, we achieve 80\% dataset coverage at 87.6\% accuracy.\vspace{-0.2cm}}
    \label{fig:jester_confidences}
\end{figure}

For completeness, we illuminate a potential SOTA integration strategy for our MovT model. We propose a cascade pipeline composed of our lightweight, inexpensive MovT architecture at the top layer and an expensive, high accuracy SOTA video understanding model at the lower layer. MovT filters a proportion of the input instances: those whose motion signature results in a confident and accurate inference. The harder, less temporal instances are then passed to the expensive lower layer. We bolster this proposal with the results presented in Fig~\ref{fig:jester_confidences}. In this study, we rank all Jester test set instances by confidence and report the performance of MovT on each test set subset above a given confidence threshold. We find that 50\% of the dataset can be accounted for at 96.0\% accuracy. Given this implementation, the cascade approach has the potential to reduce computation by almost half (given MovT's extremely lightweight nature) in comparison to solely employing an expensive SOTA video understanding model.

\vspace{-0.3cm}

Furthermore, our previous experiments have shown MovT's exceptional robustness to reductions in training data  (Fig.~\ref{fig:generalization}) and data fidelity (Fig.~\ref{fig:per-instance_reduction}) and that 5-6 point-tracks can effectively represent an entire video (Fig.~\ref{fig:importance_scores}). All of these studies suggest that our point-track pipeline has high potential to be made even more lightweight, furthering its suitability for this cascade pipeline.

\section{Efficiency Analysis}

We analyze and report the efficiency of our experimental pipeline, including preprocessing and model training. We use the Jester action recognition dataset as an example to illustrate this comparison, as it is employed in every experiment.

\subsection{Preprocessing}

To create our motion representations, we run CoTracker3 on each video instance within each dataset. Subsequently, we have produced a corresponding set of point-track trajectories for each video instance. These are then used for training and evaluation of our motion representation perception models. The Jester dataset contains 148,092 videos, each with a height of 100 pixels and varying width. As described previously, we track 60 points over 32 frames for each video. On our hardware, 8 NVIDIA GeForce RTX 2080 Ti GPUs, this preprocessing takes 3 hours.

This preparation only needs to be performed once for each dataset. This means that experimenting, training, and iterating over motion architectures is around one order of magnitude faster than using current video technology.

\subsection{Model Comparisons}

Our experiments focus on our \model and \spmodel models and the open source model, VideoMAE. Table~\ref{tab:model_comparisons} illustrates a comparison between the complexities and training times of these models.

\begin{table}[h!]
\centering
\begin{tabular}{|c||c c c|} 
 \hline
 Model & GFLOPs & Parameters & Training Time \\ [0.5ex] 
 \hline\hline
 MovT & 0.52 & 4.56M & 50 mins \\
 PixT & 0.51 & 4.41M & 45 mins \\
 \hline
 VideoMAE & * & 16.63M & 330 mins \\
 \hline
\end{tabular}
\caption{A comparison of the model complexities and training times for our different models (trained on the Jester dataset). VideoMAE is trained on full-scale images, while \model and \spmodel are trained on point-tracks and scaled-down images. *VideoMAE GFLOPs varies, see Table 1 in our main paper submission.}
\label{tab:model_comparisons}
\end{table}

The slight discrepancy between the complexities of the \model and \spmodel models can be attributed to the difference in the positional embedding method used. In all other aspects, the architectures of these models are identical.

For the sake of completeness, we present a comparison between the model complexities and performances of MovT and VideoMAE-small for equal numbers of frames below.

\begin{table}[h]
  \vspace{-0.3cm}
  \centering
  \footnotesize 
  \setlength{\tabcolsep}{3pt}
  \begin{tabular}{l cc cc}
    & \multicolumn{2}{c}{\textbf{4 Frames}} & \multicolumn{2}{c}{\textbf{16 Frames}} \\
    Model & GFLOPs ($\downarrow$) & Acc. ($\uparrow$) & GFLOPs ($\downarrow$) & Acc. ($\uparrow$) \\
    \midrule
    MovT & 0.30 & 70.86\% & 0.39 & 77.56\% \\
    VideoMAE-s & 6.49 \textbf{\textcolor{red}{($22\times$)}} & 90.64\% & 25.95 \textbf{\textcolor{red}{($67\times$)}} & 95.85\% \\
  \end{tabular}
  \vspace{-0.31cm}
\end{table}
